%% file: neurips_2024.tex
\documentclass{article}

\pdfoutput=1

\usepackage[preprint]{neurips_2024}

\usepackage{natbib}
\setcitestyle{numbers,square,comma}

\usepackage{colortbl}
\usepackage{color}
\usepackage{enumitem}
\usepackage{multirow}
\usepackage{graphicx}
\usepackage{wrapfig}
\usepackage{amsmath}
\usepackage{pifont}
\usepackage{subfigure}
\usepackage{caption}
\usepackage{subcaption}
\usepackage[utf8]{inputenc} 
\usepackage[T1]{fontenc}    
\usepackage{hyperref}       
\usepackage{url}            
\usepackage{booktabs}       
\usepackage{amsfonts}       
\usepackage{nicefrac}       
\usepackage{microtype}      
\usepackage{xcolor}         
\usepackage{bm}
\definecolor{fid}{HTML}{1280B0}
\definecolor{clip}{HTML}{bc6d4c}
\title{$\Delta$-DiT: A Training-Free Acceleration Method Tailored for Diffusion Transformers}

\author{%
Pengtao Chen$^{1\dagger}$ \quad
Mingzhu Shen$^{2\dagger}$ \quad
Peng Ye$^1$ \quad
Jianjian Cao$^1$ \quad
Chongjun Tu$^1$ \quad
\\
\textbf{Christos-Savvas Bouganis}$^2$ \quad
\textbf{Yiren Zhao}$^2$ \quad
\textbf{Tao Chen}$^1$\thanks{\vspace{-15pt}Corresponding Author.~~~$^\dagger$Equal Contribution.} \\
$^1$Fudan University \quad $^2$Imperial College London \\
{\tt\small pengt.chen@gmail.com, eetchen@fudan.edu.cn}
}

\begin{document}

\maketitle

\begin{figure*}[h]
    \centering
    \vspace{-8mm}
    \includegraphics[width=0.99 \linewidth]{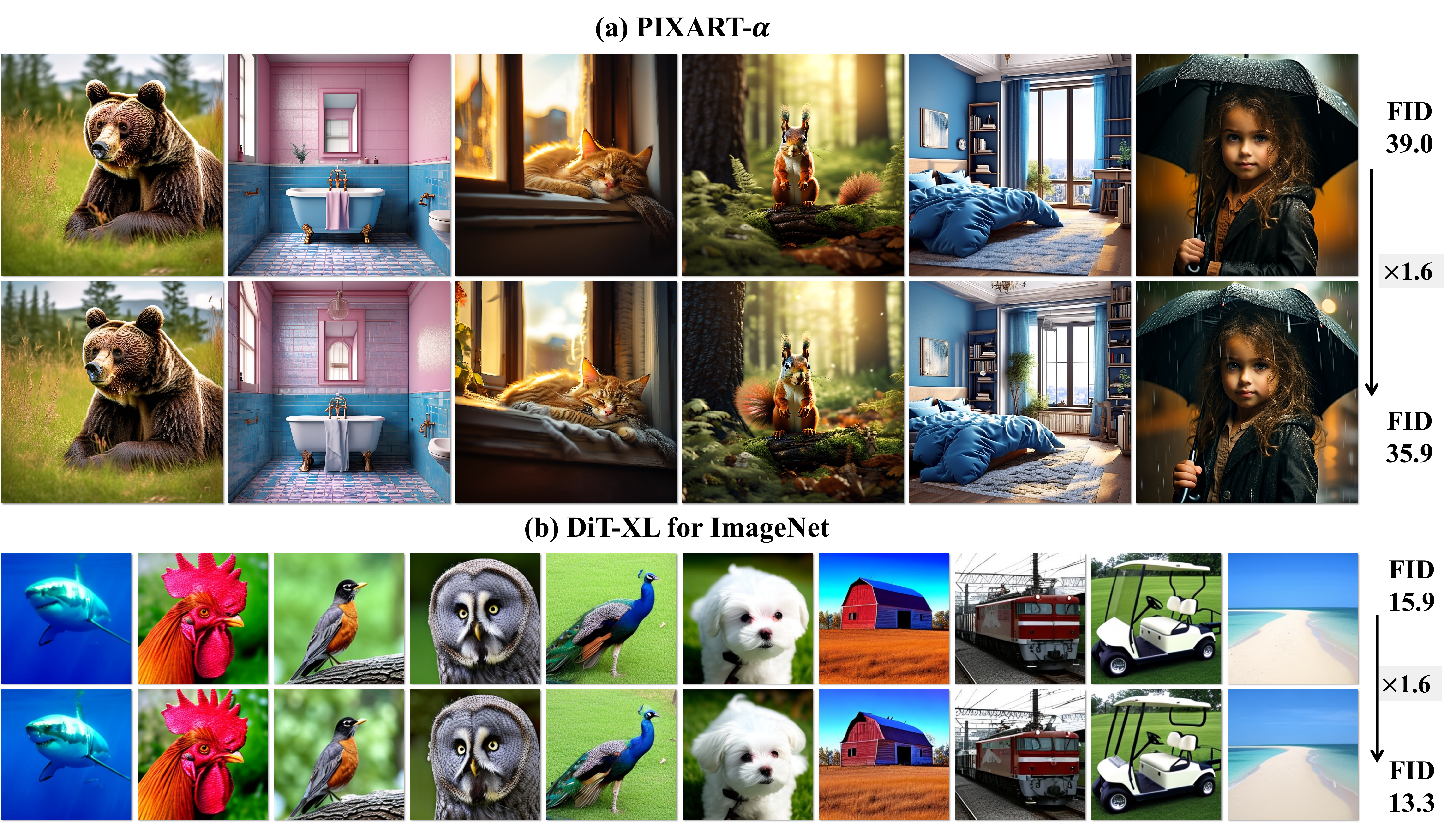}
    \vspace{-2mm}
    \caption{Accelerating diffusion transformer like PIXART-$\alpha$ and DiT-XL under the 20-step sampling.}
    
    \label{fig:previous}
\end{figure*}

\begin{abstract}
Diffusion models are widely recognized for generating high-quality and diverse images, but their poor real-time performance has led to numerous acceleration works, primarily focusing on UNet-based structures. With the more successful results achieved by diffusion transformers (DiT), there is still a lack of exploration regarding the impact of DiT structure on generation, as well as the absence of an acceleration framework tailored to the DiT architecture. 
To tackle these challenges, we conduct an investigation into the correlation between DiT blocks and image generation. Our findings reveal that the front blocks of DiT are associated with the outline of the generated images, while the rear blocks are linked to the details. 
Based on this insight, we propose an overall training-free inference acceleration framework $\Delta$-DiT: using a designed cache mechanism to accelerate the rear DiT blocks in the early sampling stages and the front DiT blocks in the later stages. 
Specifically, a DiT-specific cache mechanism called $\Delta$-Cache is proposed, which considers the inputs of the previous sampling image and reduces the bias in the inference. 
Extensive experiments on PIXART-$\alpha$ and DiT-XL demonstrate that the $\Delta$-DiT can achieve a $1.6\times$ speedup on the 20-step generation and even improves performance in most cases. 
In the scenario of 4-step consistent model generation and the more challenging $1.12\times$ acceleration, our method significantly outperforms existing methods.
Our code will be publicly available.
\end{abstract}

\input{latex/introduction}
\input{latex/related_work}
\input{latex/preliminary}

\input{latex/method}

\input{latex/experiment}

\input{latex/conclusion}

\clearpage
{\small
  \bibliographystyle{unsrt}
  \bibliography{reference}
}

\end{document}

%% file: latex/introduction.tex
\section{Introduction}

In recent years, the field of generative models has experienced rapid advancements. Among these, diffusion models~\cite{ho2020denoising, rombach2022high, yang2021score} have emerged as pivotal, attracting widespread attention for their ability to generate high-quality and diverse images~\cite{dhariwal2021diffusion}. This has also spurred the development of many meaningful applications, such as image editing~\cite{kawar2023imagic, zhang2023adding}, 3D generation~\cite{tang2023make, tang2023dreamgaussian, mo2023dit}, and video generation~\cite{wu2023tune, khachatryan2023text2video, blattmann2023stable, luo2023videofusion}. Although diffusion models have strong generation capabilities, their iterative denoising nature results in poor real-time performance.

Subsequently, numerous inference acceleration frameworks for diffusion models have been proposed. These include pruning~\cite{fang2023diffpruning, zhang2024laptop}, quantization~\cite{shang2023post, so2023tdqdm, DBLP:conf/nips/HeLLWZZ23, DBLP:conf/nips/LiX0S023}, and distillation~\cite{kim2023architectural, Tim2022progress, luo2023latent, zhao2023mobilediffusion, sauer2023adversarial} of the noise estimation network, optimizing sampling solver~\cite{song2021ddim, lu2022dpm, lu2022dpm++, zhang2023deis, karras2022elucidating}, and cache-based acceleration methods~\cite{ma2023deepcache, li2023faster}. However, almost all of these acceleration techniques are designed for the UNet-based~\cite{Olaf2015unet} architecture. Recently, Diffusion transformers (DiT)~\cite{peebles2023iccv} have achieved unprecedented success, exemplified by models like SD3.0~\cite{esser2024scaling}, PIXART-$\alpha$~\cite{chen2023pixart}, and Sora~\cite{videoworldsimulators2024}. It has surpassed current UNet models to some extent. SDXL-Lightning~\cite{lin2024sdxl} has also highlighted the redundancy of the encoder of UNet. In the current landscape where DiT has such an advantage, there's been limited inference acceleration work for DiT models. Prior work~\cite{moon2023early} introduced an early stopping strategy for DiT blocks, which requires training and is not suitable for the current context of small-step generation. For these reasons, there's an urgent need for a new acceleration framework tailored to DiT, potentially even a training-free framework.
 
However, there is still a lack of deep investigation into the DiT structure which restrains the DiT accelerating research. First, unlike traditional UNet, the DiT has a unique isotropic architecture that lacks encoders, decoders, and different depth skip connections. This causes the existing feature reuse mechanism such as DeepCache~\cite{ma2023deepcache} and Faster Diffusion~\cite{li2023faster}, may result in the loss of information when applied for DiT. Because they cache the output feature map of the block, while the DiT model without skip structure will lose the sampling input from the previous step.
Second, the impact of different components in a whole DiT structure on the generated image quality remains unexplored. DiT is composed of many blocks, these blocks are located in different depths and play different roles. For example, the front block is focused on low-level information, while the back block is focused on semantic information, but there is little research that can present a comprehensive qualitative and quantitative analysis for these blocks. This further causes us to be uncertain about blocks to target when accelerating the DiT network. 

For the first challenge, we propose the $\Delta$-Cache method, which involves using the offset of features rather than the feature maps themselves as cache objects to avoid the loss of the input information. Regarding the second challenge, we have discovered that the transformer blocks in the front part of DiT are more relevant for generating image outlines, while those in the later part are more relevant for generating image details. And combining previous research~\cite{wang2023diffusion, liu2023oms, Amir2023iclr}, which suggests that diffusion models generate outlines in the early stages of sampling and details in the later stages, we propose a stage-adaptive inference acceleration method ($\Delta$-DiT) that aligns with this sampling characteristic. Specifically, we $\Delta$-Cache the blocks in the rear part of DiT for approximation during early-stage sampling and $\Delta$-Cache the blocks in the front part during later-stage sampling. We evaluated our method on multiple datasets, including MS-COCO2017~\cite{lin2014microsoft} and PartiPrompts~\cite{partiprompt2022}, and conducted experiments on various DiT architecture models such as PIXART-$\alpha$~\cite{chen2023pixart}, DiT-XL~\cite{peebles2023iccv}, and PIXART-$\alpha$-LCM~\cite{chen2023pixart, luo2023latent}. Extensive quantitative results demonstrate the effectiveness of our method. Specifically, based on the 20-step generation, we achieved a 1.6x speedup with results that are comparable to or better than the baseline (FID: 39.002$\to$35.882). In the more challenging 4-step generation scenarios, our method significantly outperformed existing baseline methods at more extreme acceleration ratios (1.12$\times$). 

\vspace{-3pt}
The contributions of our paper are as follows:
\vspace{-4.5pt}
\begin{itemize}[leftmargin=1em]
\item We first discovered that the blocks at the front of DiT focus more on generating image outlines, while the blocks at the back concentrate on details. Similarly, diffusion models generate outlines in the early stages and details in the later stages. These insights inspired the design of $\Delta$-DiT.
\vspace{-1pt}
\item We propose the first inference acceleration method $\Delta$-DiT without training specifically designed for diffusion transformers, which achieves the acceleration by caching the back block of DiT in the early sampling stage and the front block in the later stage.
\vspace{-0.5pt}
\item $\Delta$-DiT relies on our $\Delta$-Cache, a specialized cache mechanism designed for DiT, which effectively avoids losing information from the input (the sampled result of the previous step).
\item In the experiments: $\Delta$-DiT achieves a $1.6\times$ speedup in 20-step generation and achieve even better generation. In the scenario of 4-step consistent model generation and the more challenging $1.12\times$ acceleration, our method significantly outperforms existing methods.
\end{itemize} 

%% file: latex/related_work.tex
\section{Related Work}
\textbf{Efficient Diffusion Model.}\
To address the problem of poor real-time performance in diffusion models, various lightweight and acceleration techniques have emerged. Currently, methods for accelerating diffusion models for image generation can be broadly categorized into three perspectives:
Firstly, lightweight a noise estimation network is one approach. Similar to traditional network lightweight, many efforts focus on pruning~\cite{fang2023diffpruning, zhang2024laptop}, quantization~\cite{shang2023post, so2023tdqdm, DBLP:conf/nips/HeLLWZZ23, DBLP:conf/nips/LiX0S023}, and distillation~\cite{kim2023architectural, Tim2022progress, luo2023latent, zhao2023mobilediffusion, sauer2023adversarial} of noise estimation networks to obtain a smaller yet comparable performance model.
Secondly, optimizing the sampling time steps is a unique dimension for diffusion models. Most methods currently focus on exploring efficient ODE solvers~\cite{song2021ddim, lu2022dpm, lu2022dpm++, zhang2023deis, karras2022elucidating}, aiming to obtain high-quality samples with fewer sampling steps. Another approach attempts to optimize sampling time steps from the perspective of skips~\cite{wang2023kdd}.
Lastly, there's a focus on jointly optimizing noise estimation networks and time steps. Such methods often achieve higher acceleration ratios. For instance, OMS-DPM~\cite{liu2023oms} and Autodiffusion~\cite{li2023autodiffusion} simultaneously optimize skips and allocate noise estimation networks of specific sizes for each time step. LCM~\cite{luo2023latent} organizes noise estimation networks from the time step perspective to enable the network to generate samples with fewer steps. Approaches like deepcache~\cite{ma2023deepcache} and faster diffusion~\cite{li2023faster} consider information between steps, utilizing a cache mechanism to indirectly modify the network structure for acceleration.
However, most of the aforementioned work is implemented and validated on the UNet architecture. Only previous work~\cite{moon2023early} proposed an early stopping strategy for DiT, which significantly impacts generation results in the current focus on generating with fewer steps, and has not explored the impact of DiT block on generation. Starting from the third acceleration perspective, this paper explores the impact of DiT blocks on generation and proposes a dedicated acceleration framework urgently needed for the DiT architecture. 

\textbf{Cache Mechanism.}\
Cache mechanism is a concept in computer systems that involves temporarily storing information that may be reused in the future to improve processing speed. In large language models, the KV cache~\cite{zhenyu2023kv, suyu2023kv} method is widely used, which caches the K and V in attention computations for reuse in the next attention calculation, thereby accelerating inference. In diffusion model generation, there are also some methods based on cache principles. Deepcache~\cite{ma2023deepcache} expedites computation across adjacent steps in UNet networks by caching the output feature maps of up-sampling blocks. Faster Diffusion~\cite{li2023faster} achieves acceleration in neighboring steps of UNet networks by caching the output of the UNet encoder and the feature maps at skip connections. Meanwhile, TGATE~\cite{zhang2024cross} accelerates the later stages of sampling by caching the output feature maps of the cross-attention module. However, these methods all cache feature maps, which are not suitable for the isotropic architecture of DiT. Directly caching block output feature maps, as in Deepcache and Faster Diffusion, would lose the previous step sampled image information in DiT, which lacks skip connections. Caching only the output feature maps of more fine-grained modules, as in TGATE, would bring limited acceleration. Therefore, this paper proposes a method called $\Delta$-Cache, which caches feature offsets, to address these issues encountered in DiT.

%% file: latex/preliminary.tex
\section{Preliminary}
The concept of diffusion originates from a branch of non-equilibrium thermodynamics~\cite{de2013non} in physics. In recent years, researchers have applied this concept to image generation~\cite{ho2020denoising, rombach2022high, yang2021score, dhariwal2021diffusion, song2019generative}, transforming the process into two stages: noise diffusion and denoising.

\textbf{Noise Diffusion Stage.}\
This is also the training phase of the diffusion model. Given an original image $\bm{x_0}$ and a random time step $t\in [1, T]$ (where $T$ is the total steps), the image after $t$ steps of diffusion is $\sqrt{\bar{\alpha}_t}\bm{x_0}+\sqrt{1-\bar{\alpha}_t}\bm{\epsilon}$, where $\bar{\alpha}_t$ is constant related to $t$. The noise estimation network is then used to estimate the noise in the diffused result, making the estimated noise $\bm{\epsilon_{\theta}}$ as close as possible to the actual noise $\bm{\epsilon}$ added during diffusion. The specific gradient representation is as follows~\cite{ho2020denoising}:
\begin{equation}
    \mathcal{L}(\bm{\theta}) = \mathbb{E}_{t, \bm{x_0}\sim q(\bm{x}), \bm{\epsilon} \sim \mathcal{N}(0,1)}\left[ \|\bm{\epsilon} - \bm{\epsilon}_{\bm{\theta}}(\sqrt{\bar{\alpha}_t}\bm{x_0} + \sqrt{1-\bar{\alpha}_t}\bm{\epsilon}, t)\|^2 \right],
    \label{eq:forward}
\end{equation}
where $q(\bm{x})$ is the dataset distribution, and $\mathcal{N}$ is the Gaussian distribution. In most current works, the noise estimation networks are mostly based on UNet architecture. However, in isotropic architectures like DiT, $\bm{\epsilon_\theta}(\bm{x}_t)$ can be further transformed into $\bm{f}_{N_b}(\bm{f}_{N_b-1}(\cdots(\bm{f}_1(\bm{x}_t)))=\bm{f}_{N_b}\circ\bm{f}_{N_b-1}\circ\cdots\circ\bm{f}_1(\bm{x}_t)=\bm{F}^{N_b}_1(\bm{x}_t)$, where $\bm{f}_n$ represents the mapping of the $n$-th DiT block, and $\bm{F}_1^{N_b}$ represents the mapping of the first to the $N_b$-th DiT blocks. In the current DiT framework, the value of $N_b$ is 28.

\textbf{Denoising Stage.}\
Also known as the inference or reverse phase. This is the process from Gaussian noise to a generated image, which is also the goal of acceleration in this paper. Initially, a random Gaussian noise $\bm{x}_T$ is given. It is input into the noise estimation network $\bm{\epsilon_\theta}$ to obtain the noise estimate $\bm{\epsilon_\theta}(\bm{x}_T)$. According to specific sampling solvers, the noisy image is denoised to produce the denoised sample $\bm{x}_{T-1}$ after one step. After iterating this process $T$ times, the final generated image is obtained. Using DDPM~\cite{ho2020denoising} as an example, the iterative denoising process is as follows:
\begin{equation}
\bm{x}_{t-1} = \frac{1}{\sqrt{\alpha_t}} \left( \bm{x}_t - \frac{\beta_t}{\sqrt{1-\bar{\alpha}_t}} \bm{\epsilon}_{\bm{\theta}}(\bm{x}_t, t) \right) + \sigma_t \bm{z},    \label{eq:denoising}
\end{equation}
where $\alpha_t$, $\beta_t$ and $\sigma_t$ is constant related to $t$, and $\bm{z}\sim \mathcal{N}(\bm{0},\bm{I})$. For other solvers~\cite{song2021ddim, lu2022dpm, lu2022dpm++, zhang2023deis, karras2022elucidating}, the sampling formula differs slightly from Eq.~\ref{eq:denoising}, but they are all functions of $\bm{x}_t$ and $\bm{\epsilon_{\theta}}$. In many scenarios, the noise estimation network $\bm{\epsilon_\theta}(\bm{x}_t, t, \bm{c})$ has another input $\bm{c}$. It is conditional control information, which can be either a class embedding or a text embedding.

%% file: latex/method.tex
\section{Stage-adaptive Inference Acceleration for Diffusion Transformers}
This section will introduce our stage-adaptive inference acceleration method employed in the diffusion transformer, which is a training-free approach. In the following, we first analyze the challenges of information reuse in the current DiT model and propose a cache method specifically designed for DiT, called $\Delta$-Cache. Secondly, based on this cache method, we explore the specific effects of different parts of blocks on generation. Finally, combining these specific effects with the characteristics of diffusion generation, we propose a stage-adaptive method, $\Delta$-DiT, for accelerating DiT generation.

\subsection{Tailored Cache Method for DiT}
Feature reuse is an important strategy in training-free inference acceleration, with cache methods being prominent. Recently, cache methods have made significant strides in accelerating inference for diffusion models. However, these methods primarily focus on UNet architectures, such as DeepCache~\cite{ma2023deepcache}, Faster Diffusion~\cite{li2023faster}, and TGATE~\cite{zhang2024cross}. In this section, we will explain the challenges these methods face when applied to DiT and introduce our $\Delta$-Cache method, specifically designed to address these challenges in DiT structures.

\begin{figure*}[t]
    \centering
    \includegraphics[width=1 \linewidth]{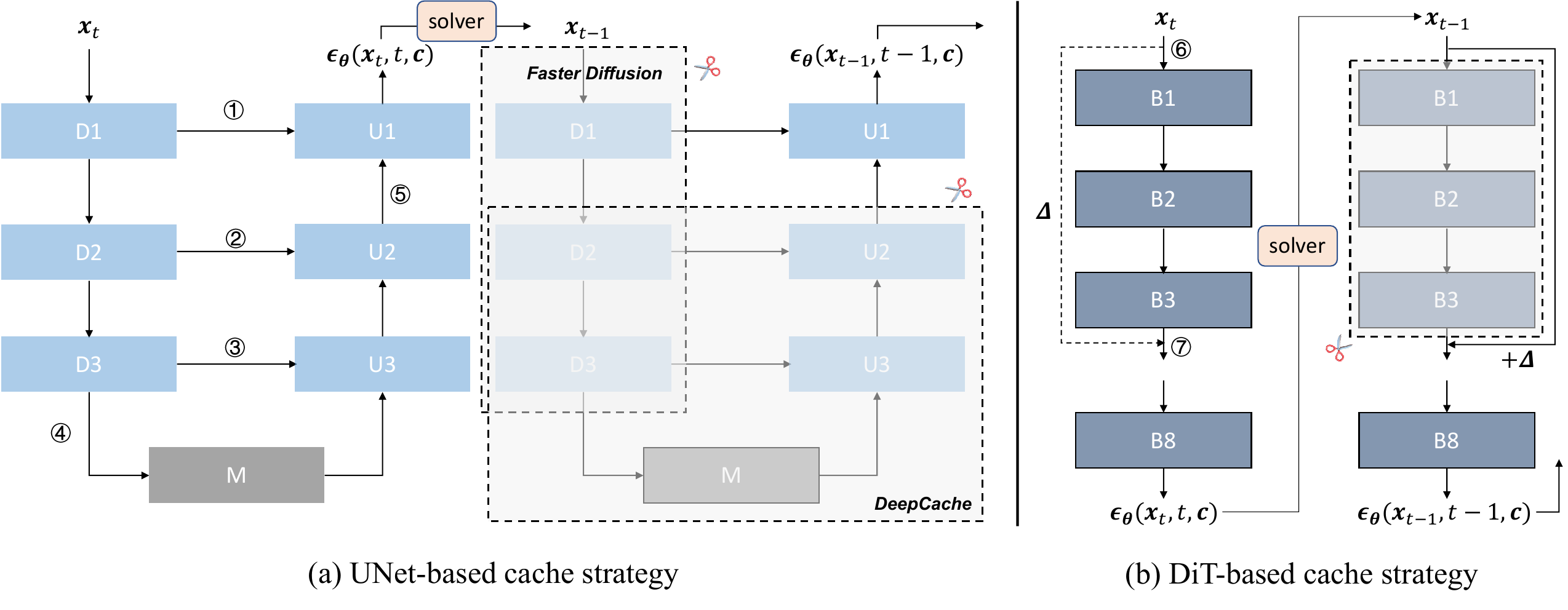}
    \vspace{-5mm}
    \caption{(a) Existing cache strategies based on the UNet architecture, such as DeepCache~\cite{ma2023deepcache} and Faster Diffusion~\cite{li2023faster}. (b) Cache strategy based on the DiT structure, where the dashed line represents the difference between two feature maps instead of the true residual structure. }
    \vspace{-5mm}
    \label{fig:cache}
\end{figure*}

\textbf{Challenges.} Figure~\ref{fig:cache}a illustrates the denoising process based on a traditional UNet over two consecutive steps ($t$ and $t-1$). The entire UNet consists of downsampling modules D, upsampling modules U, middle module M, and skip connections. The cache method in the diffusion model is to save some intermediate feature maps of the previous UNet and reuse them for speedup in the next UNet. For the Faster Diffusion~\cite{li2023faster}, the cache positions are \ding{172}-\ding{175}, which means that during the next step, the computations for D1-D3 are skipped. DeepCache~\cite{ma2023deepcache}, on the other hand, caches at position \ding{176}, skipping D2, D3, M, U3, and U2. TGATE~\cite{zhang2024cross} caches the output feature maps of the cross-attention modules within D1-D3 and U1-U3, providing less benefit due to its finer granularity, which is not discussed in detail here. Comparing the first two methods, we can see that the Faster Diffusion skips the encoder (D1-D3), thus losing the input from the previous sampling step $\bm{x}_{t-1}$. In contrast, DeepCache only skips the deep network structures, so the previous sampling result can still supervise the output via the D1-skip-U1 path. However, in the DiT framework, these methods essentially converge into a single approach. Both DeepCache and Faster Diffusion focus on caching the feature maps output by specific blocks. Figure~\ref{fig:cache}b depicts a is isotropic network composed of a series of DiT blocks. If we also use the output feature map of a specific block in DiT (assume at position \ding{178}) as the cache target, we find that this approach also loses information from the previous sampling step $\bm{x}_{t-1}$ because it directly skips the computation before B3. This issue is similar to the problem encountered in Faster Diffusion, so we refer to this cache method in DiT as the DiT version of Faster Diffusion. This method, which lacks information from the previous samples, has disadvantages in generating images and can be particularly catastrophic in scenarios with a small number of generation steps.

\textbf{Method.} To address the problem of missing the previous sampling result due to caching feature maps in DiT, this paper proposes a novel cache mechanism: $\Delta$-Cache. Instead of the traditional approach of caching feature maps, $\Delta$-Cache caches the deviations between feature maps. As illustrated in Figure~\ref{fig:cache}b, we cache the difference between feature maps at position \ding{177} and \ding{178} rather than the feature map at position \ding{178}. This allows us to skip the computation of B1-B3 in the next step, while the sampling result of the previous step $\bm{x}_{t-1}$ can still be incorporated into the output of the latter step via the virtual construction $\Delta$. Mathematically, by caching $\bm{F}_{1}^{N_c}(\bm{x}_{t})-\bm{x}_{t}$, we can skip the first $N_c$ blocks in DiT, which forms the initial version of the $\Delta$-Cache tailored for DiT. Further, we can find that $\Delta$-Cache has the following three advantages: (a) It resolves the issue of degraded generation quality caused by losing previous sampling information when caching the output feature maps of DiT blocks. (b) Unlike Faster Diffusion in DiT, which can only skip the first few blocks, $\Delta$-Cache allows skipping of front, middle, or back blocks, offering greater flexibility. (c) It is well-suited for isotropic architecture, such as DiT, where the output feature map scale of each block is consistent, thus enabling the computation of $\Delta$. In the experimental section, we will demonstrate the effectiveness of this method for DiT.

\subsection{Effect of DiT Blocks on Generation}
\label{effectdit}

\begin{wrapfigure}{r}{0.5\textwidth}
    \vspace{-25pt}
    \centering
    \includegraphics[width=1 \linewidth]{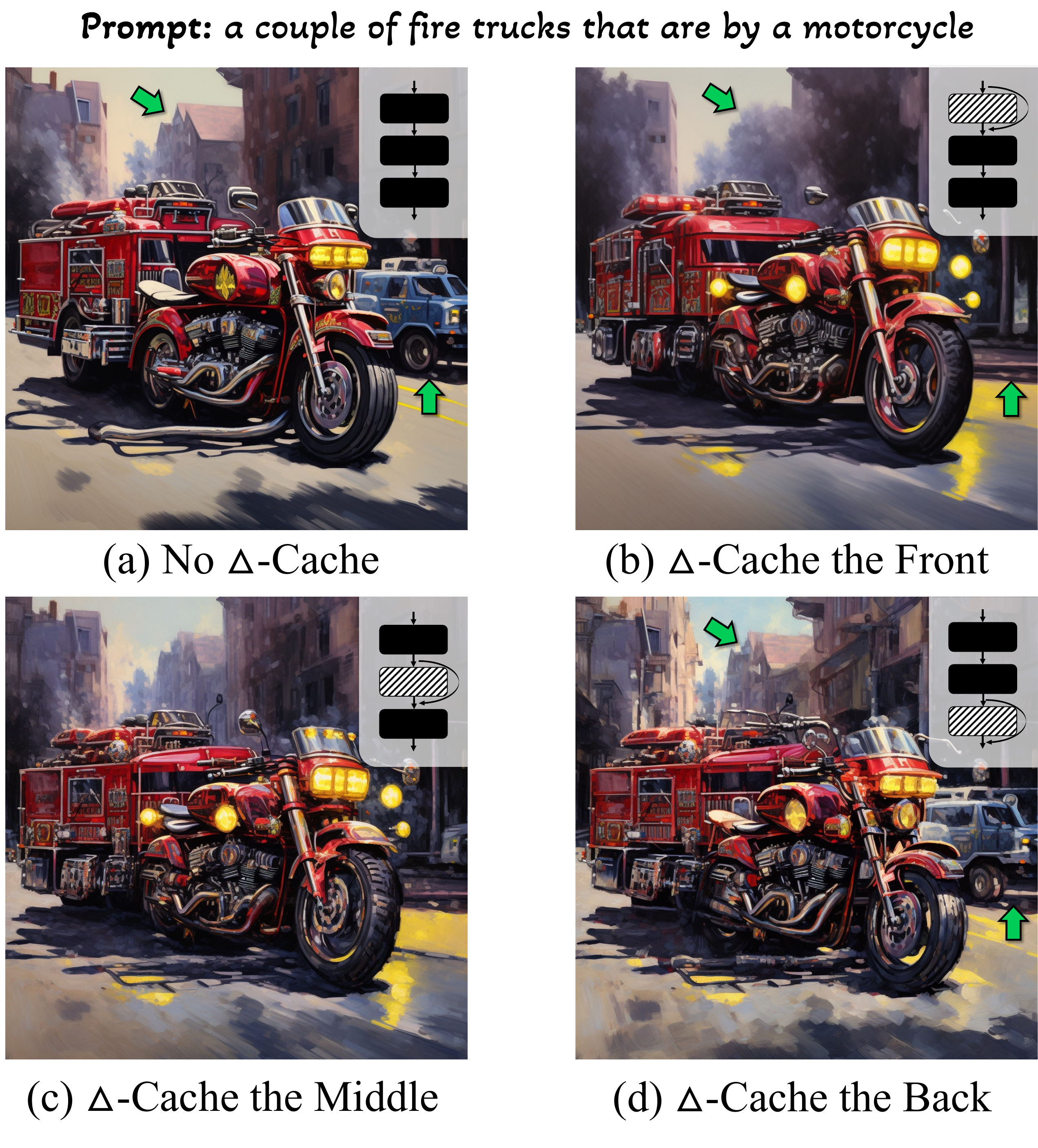}
    \vspace{-20pt}
    \caption{Images generated by $\Delta$-Cache for different parts of blocks in DiT.}
    \label{fig:motivation}
    \vspace{-12pt}
\end{wrapfigure}
\label{method: delta-cache}
In the previous part, we proposed a cache mechanism for the DiT architecture called $\Delta$-Cache, which is flexible and can be applied to different parts of the DiT blocks. However, current research lacks exploration into the specific impact of different parts of DiT blocks on the final generated image, leading to uncertainty about which blocks should be targeted by $\Delta$-Cache. Therefore, in this part, we will conduct the first exploration of this gap using $\Delta$-Cache.

\textbf{Qualitative.} DiT architectures have a large number of blocks, and classical models of DiT architectures like DiT-XL and PIXART-$\alpha$ contain 28 blocks per network. So in the following, we will explore the impact of these blocks on the generated results from a coarse-grained perspective. For clarity, we designate the initial block position that requires $\Delta$-Cache as $I$, and denote the number of blocks to cache as $N_c$. In the exploratory experiments, we conducted experiments with $I=$$1$, $4$, $7$, and $N_c=21$, representing $\Delta$-Cache on the front, middle, and back blocks of DiT, respectively. Due to the relatively small number of generation steps, the cache interval is set to $N=2$, meaning cache occurs every 2 steps. Based on these settings, we obtain the images for the different configurations, which are shown in Figure~\ref{fig:motivation}. By comparing the images generated without cache and those generated with various $\Delta$-Cache objects, several observations can be made. Caching the front blocks results in less accurate outline generation. For example, in Figure~\ref{fig:motivation}a, the outlines of the blue car and roof on the right side are clear, whereas in Figure~\ref{fig:motivation}b, these outlines are lost, resulting in a smoother overall image appearance. Conversely, caching the back blocks preserves the outlines better but produces poorer details. In Figure~\ref{fig:motivation}d, the outlines of the blue car and roof are retained, but there is a lack of pixel-level details. Caching the middle blocks offers a compromise, with better detail generation compared to only caching the back blocks and better outline generation compared to only caching the front blocks. 

\begin{wraptable}[9]{r}{0.6\textwidth}
\vspace{-12pt}
\centering
\small
\caption{Quantitative results of caching different blocks in PIXART-$\alpha$ on 500 MS-COCO2017 samples.}
\resizebox{\linewidth}{!}{ 
\begin{tabular}{l|cc|c}
\toprule
      \multirow{2}*{\bf Cache Object} & \multicolumn{2}{c|}{\bf Outline}&\bf Detail\\
     & \bf Ave. Gradient $\uparrow$& \bf High-freq Error $\downarrow$& \bf PIQE $\downarrow$\\
\midrule 
 Front Blocks& 36.170& 0.866& \bf 13.137\\
 Middle Blocks& 42.836& 0.838& 15.219\\
 Back Blocks& \bf 59.320& \bf 0.776& 19.974\\
\bottomrule
\end{tabular}}
\vspace{-10pt}
\label{table:motivation}
\end{wraptable}

\textbf{Quantitative.} The above is just a demonstration with a single prompt. To validate our findings, we conducted quantitative verification on a subset of MS-COCO2017. To quantitatively characterize the ability to generate outlines, we used the average gradient of images based on the Sobel operator and the high-frequency loss based on the Discrete Fourier Transform~\cite{yang2023diffusion}. To quantitatively assess the detail generation ability of the model, we used the classical blind image quality assessment metric PIQE~\cite{venkatanath2015blind}, which effectively captures blockiness and is well-suited for this scenario. The quantitative results are shown in Table~\ref{table:motivation}. Caching the back blocks (i.e., only calculating the front blocks) results in lower high-frequency loss and higher average gradient, indicating better performance in outline generation. Conversely, caching the front blocks results in higher high-frequency loss, lower average gradient, and higher PIQE scores, indicating poorer local block generation and greater distortion. This further confirms that the front blocks in DiT are more related to outline generation, while the back blocks are more related to detail generation.

\begin{figure*}[b]
    \small
    \centering
    \vspace{-4mm}
    \includegraphics[width=1\linewidth]{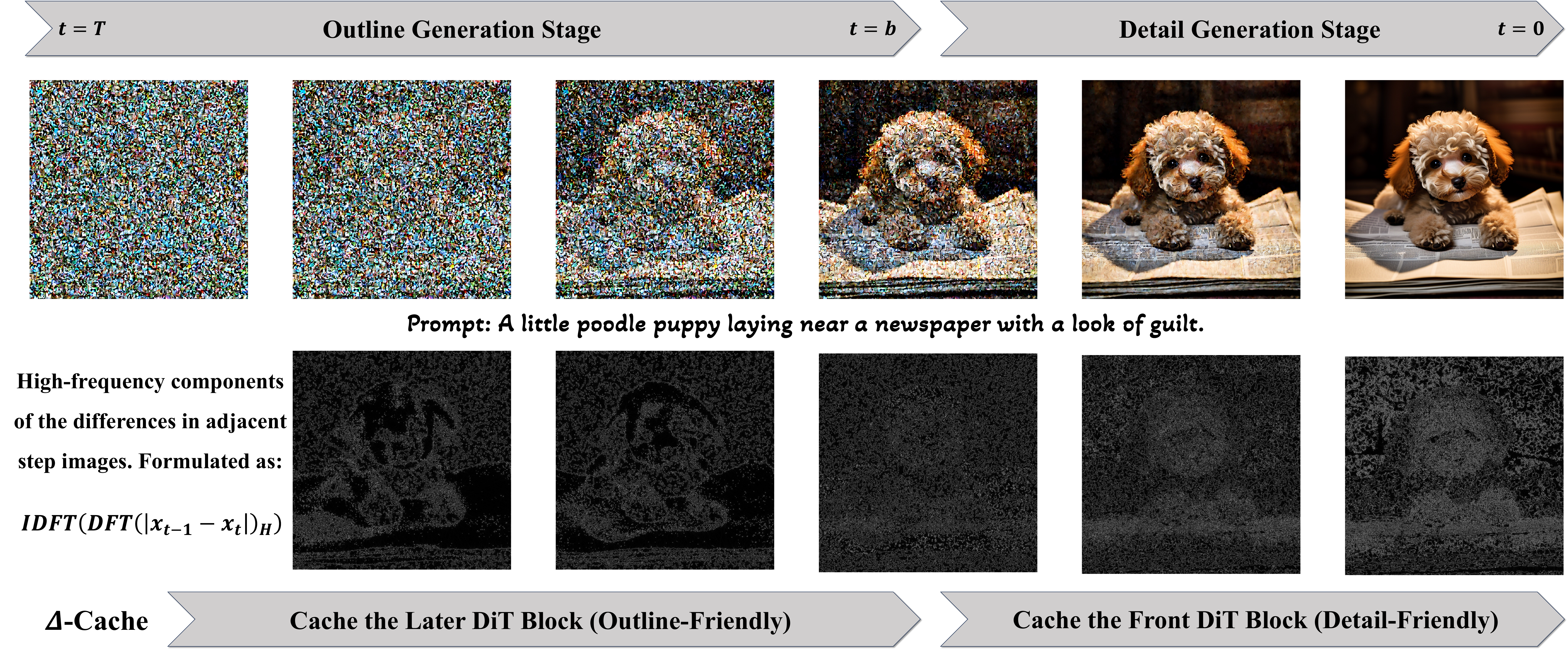}
    \caption{\textbf{Overview of the $\Delta$-DiT}: The diffusion model emphasizes generating outlines early in sampling and details later. Our previously proposed $\Delta$-Cache method caches back blocks for outline-friendly generation and front blocks for detail-friendly results. In the $\Delta$-DiT, the properties of the diffusion model and $\Delta$-Cache are aligned in stages, that is, $\Delta$-Cache is applied to the back blocks in the DiT during the early outline generation stage of the diffusion model, and on front blocks during the detail generation stage. The stage is bounded by a hyperparameter $b$.}
    \label{fig:method}
\end{figure*}

\subsection{Stage-adaptive Acceleration Method for DiT}

After addressing cache bottlenecks in DiT and analyzing the impact of DiT blocks on generation, this part proposes $\Delta$-DiT, an overall training-free inference acceleration method for the DiT architecture. It combines the above findings with the sampling characteristics of diffusion. Current research~\cite{wang2023diffusion, liu2023oms, Amir2023iclr} indicates that the diffusion inference process has a characteristic: in the early stages of the denoising process, diffusion models focus on generating outlines, while in the later stages, they focus more on generating details. In Section~\ref{effectdit}, we found that caching the back DiT blocks is favorable for outline generation while caching the front DiT blocks is favorable for detail generation. Leveraging these two findings, we propose a stage-adaptive acceleration method called $\Delta$-DiT. The specific process is illustrated in Figure~\ref{fig:method}. From left to right, it shows the denoising process of the diffusion model (from $t=T$ to $t=0$). The upper part of the image represents the denoising results at different timesteps, while the lower part shows the high-frequency components of the differences between adjacent timestep images. Specifically, these high-frequency components are obtained by transforming the image into the frequency domain using a Discrete Fourier Transform (DFT), then isolating the high-frequency components and transforming them back to the time domain using an Inverse Discrete Fourier Transform (IDFT). In the early sampling stage, high-frequency signals mainly focus on the outlines, while in the later sampling stage, they mainly focus on details. These high-frequency signals represent the areas the model focuses on at each timestep. Since the diffusion model focuses on outline generation in the early stage and the $\Delta$-Cache method for the back block is outline generation friendly, $\Delta$-Cache is applied to the back block in the outline generation stage. On the contrary, since the diffusion model focuses on details generation in the later stage, the $\Delta$-Cache of the front blocks are detail generation friendly, so the $\Delta$-Cache is applied to the front blocks in the detail generation stage.

In this training-free process, there is a hyperparameter denoted as $b$, which represents the boundary between the outline generation stage and the detail generation stage. When $t\leq b$, $\Delta$-Cache is applied to the back blocks; when $t>b$, $\Delta$-Cache is applied to the front blocks. The number of blocks that require $\Delta$-Cache will be determined based on the actual computational requirements. Suppose the computation cost of one block is $M_b$ and the expected total computation cost is $M_g$. As previously mentioned, the cache interval is $N$, and the number of DiT blocks is $N_b$. First, we roughly determine the value of $N = \lceil \frac{T\times N_b\times M_b}{M_g} \rceil$. In some current low-step scenarios, the value of $N$ is often set to 2. After determining $N$, the actual number of blocks to cache at the timestep is:
\begin{equation}
  \begin{aligned}
    N_c = [(\underbrace{\vphantom{\frac{M_g-(T \bmod N)\times N_b \times M_b}{\lfloor T/N \rfloor \times M_b}}\frac{M_g-(T \bmod N)\times N_b \times M_b}{\lfloor T/N \rfloor \times M_b}}_{\text{the computation in each $N$ step}}-\underbrace{\vphantom{\frac{M_g-(T \bmod N)\times N_b \times M_b}{\lfloor T/N \rfloor \times M_b}}N_b\times M_b}_{\text{the first full DiT}}) / \underbrace{\vphantom{\frac{M_g-(T \bmod N)\times N_b \times M_b}{\lfloor T/N \rfloor \times M_b}}(M_b\times(N-1))}_{\text{the remaining cached steps}}]
  \end{aligned} \label{eq:method}
\end{equation}
Although the actual computation cost cannot perfectly match the ideal computation cost, they are very close and generally meet the requirements. Once these parameters are determined, the inference process becomes fixed, enabling acceleration without the need for further training.

%% file: latex/experiment.tex
\section{Experiment}

\subsection{Experimental Settings}
\textbf{Models, Evaluation Data and Solvers.}
We conduct experiments on three diffusion transformer-based architectures, DiT-XL~\cite{peebles2023iccv}, PIXART-$\alpha$~\cite{chen2023pixart} and PIXART-$\alpha$-LCM~\cite{chen2023pixart, luo2023latent}.
For DiT-XL, we use 1000 classes from ImageNet~\cite{olga2015imagenet} to generate 50k images for evaluation. For the PIXART-$\alpha$ series models, we use 1.632k prompts from Partipromt~\cite{partiprompt2022} and 5k prompts from the validation dataset of MS-COCO2017~\cite{lin2014microsoft} to evaluate the quality of generated images. For consistency, in our main experiment, we default to using the 20-step DPMSolver++~\cite{lu2022dpm++}, which is the default setting of PIXART-$\alpha$. For consistency model generation, we use the 4-step LCMSolver~\cite{song2023consistency}. In the ablation study, we will compare more advanced solvers such as DEIS~\cite{zhang2023deis} and EulerD~\cite{karras2022elucidating}.

\textbf{Baseline.}\
To demonstrate the effectiveness of our method, we compared it with several existing fast-generation methods as well as some techniques adapted from UNet acceleration. These methods include the generation methods of the base models PIXART-$\alpha$~\cite{chen2023pixart}, DiT-XL~\cite{peebles2023iccv}, and PIXART-$\alpha$-LCM~\cite{chen2023pixart}, the DiT acceleration strategy based on the Faster Diffusion~\cite{li2023faster} concept, the TGATE~\cite{zhang2024cross}, and the $\Delta$-Cache proposed for different parts of the block.

\textbf{Evaluation Metrics.}
We use a range of metrics to evaluate the generation efficiency and image quality. 
To evaluate the generation efficiency, the MACs, and latency are adopted to measure the theoretical computational complexity and the practical time consumed to generate an image, respectively. Lower MACs and latency mean higher generation efficiency. And the speed is the acceleration rate. For generation quality, we choose the widely applied FID~\cite{heusel2017gans}, IS~\cite{clipscore2021}, and CLIP-Score~\cite{salimans2016improved}.

\begin{table}[t]
\centering
\small
\caption{Results on the PIXART-$\alpha$. Gate is the hyperparameter mentioned in the paper of TGATE~\cite{zhang2024cross}. Latency is measured in milliseconds and was evaluated on an A100.}
\resizebox{\linewidth}{!}{ 
\begin{tabular}{l|ccc|ccc|c}
\toprule
     \multirow{2}*{\bf Method} & \multirow{2}*{\bf MACs $\downarrow$} & \multirow{2}*{\bf Speed $\uparrow$}&\multirow{2}*{\bf Latency $\downarrow$}& \multicolumn{3}{c|}{\bf MS-COCO2017} & \bf PartiPrompts \\
     &  &   && \bf FID $\downarrow$ & \bf IS $\uparrow$ & \bf CLIP $\uparrow$ & \bf CLIP $\uparrow$ \\
\midrule
         PIXART-$\alpha$ ($T=20$)~\cite{chen2023pixart}&  85.651T&  1.00$\times$&2290.668&  39.002&  31.385&  \textbf{30.417}& \underline{30.097}\\
         PIXART-$\alpha$ ($T=13$)~\cite{chen2023pixart}&  55.673T&  1.54$\times$&1565.175&  39.989&  30.822&  \underline{30.399}& 29.993\\
\midrule 
         
         Faster Diffusion ($I=14$)~\cite{li2023faster}&  64.238T&   1.33$\times$&1777.144&  41.560&  31.233&  30.300& 29.958\\
         Faster Diffusion ($I=21$)~\cite{li2023faster}&  53.532T&   1.60$\times$&1517.698&  42.763&  30.316&  30.227& 29.922\\
\midrule 
         TGATE (Gate=10)~\cite{zhang2024cross}&  61.075T&   1.40$\times$&1718.308&  37.413&  31.079&  29.782& 29.347\\
         TGATE (Gate=8)~\cite{zhang2024cross}&  56.170T&   1.52$\times$&1603.250&  37.539&  30.124&  29.021& 28.654\\
\midrule 
 $\Delta$-Cache (Front Blocks)& 53.532T&  1.60$\times$&1522.346& 41.702& 30.276& 30.288&29.964\\
 $\Delta$-Cache (Middle Blocks)& 53.532T&  1.60$\times$&1522.528& \underline{35.907}& \textbf{33.063}& 30.183&30.078\\
 $\Delta$-Cache (Last Blocks)& 53.532T&  1.60$\times$&1522.669& \textbf{34.819}& \underline{32.736}& 29.898&\underline{30.099}\\
\midrule 
 \rowcolor[HTML]{EFEFEF} 
 Ours ($b=12$)& 53.532T&  1.60$\times$&1534.551& \underline{35.882}& \underline{32.222}& \underline{30.404}&\textbf{30.123}\\
\bottomrule
\end{tabular}}
\vspace{-6mm}
\label{table:pixart-20}
\end{table}

\begin{table}[t]
\centering
\small
\caption{Results on the PIXART-$\alpha$-LCM. The default number of generation steps $T$ is 4. }
\resizebox{\linewidth}{!}{ 
\begin{tabular}{l|ccc|ccc|c}
\toprule
     \multirow{2}*{\bf Method} & \multirow{2}*{\bf MACs $\downarrow$} & \multirow{2}*{\bf Speed $\uparrow$}&\multirow{2}*{\bf Latency $\downarrow$}& \multicolumn{3}{c|}{\bf MS-COCO2017} & \bf PartiPrompts \\
     &  &   && \bf FID $\downarrow$ & \bf IS $\uparrow$ & \bf CLIP $\uparrow$ & \bf CLIP $\uparrow$ \\
\midrule
         PIXART-$\alpha$-LCM~\cite{chen2023pixart}&  8.565T&  1.00$\times$
&415.255&  40.433&  30.447&  29.989& 29.669\\
\midrule
         Faster Diffusion ($I=4$)~\cite{li2023faster}&  7.953T&   1.08$\times$
&401.137&  468.772&  1.146&  -1.738& 1.067\\
         Faster Diffusion ($I=6$)~\cite{li2023faster}&  7.647T&   1.12$\times$
&391.081&  468.471&  1.146&  -1.746& 1.057\\
\midrule 
         TGATE (Gate=2)~\cite{zhang2024cross}&  7.936T&   1.08$\times$
&400.256&  42.038&  29.683&  29.908& 29.549\\
         TGATE (Gate=1)~\cite{zhang2024cross}&  7.623T&   1.12$\times$
&398.124&  44.198&  27.865&  29.074& 28.684\\
\midrule
 \rowcolor[HTML]{EFEFEF} 
 Ours ($b=2$, $N_c=4$)& 7.953T&  1.08$\times$&400.132& 39.967& 29.667& 29.751&29.449\\
 \rowcolor[HTML]{EFEFEF} 
 Ours ($b=2$, $N_c=6$)& 7.647T&  1.12$\times$&393.469& 40.118& 29.177& 29.332&29.226\\
\bottomrule
\end{tabular}}
\vspace{-5mm}
\label{table:pixart-lcm-4}
\end{table}

\subsection{Comparison with the Baseline Model}
We provide a comprehensive comparison with fast generation methods for PIXART-$\alpha$ on both the generation efficiency and image quality in Table~\ref{table:pixart-20}. The proposed method exceeds the baseline PIXART-$\alpha$ on all metrics except for a small gap in the MS-COCO2017 CLIP-Score, with a 1.60$\times$ speedup. When the inference costs are aligned, we surpass PIXART-$\alpha$ on all metrics by a large margin (e.g., FID: 39.989 $\to$ 35.882). 
Moreover, our proposed method also outperforms Faster Diffusion and TGATE in all metrics on both datasets with similar or even higher generation efficiency.
In Table~\ref{table:dit-20}, we demonstrate the efficiency and effectiveness of our proposed method for the DiT-XL architecture.
\begin{wraptable}{r}{0.7\textwidth}
\small
\centering
\vspace{-2mm}
\caption{Results on the DiT-XL. Because the TGATE can only handle cross-attention, it cannot be used for DiT-XL.}
\vspace{-2mm}
\resizebox{\linewidth}{!}{ 
\begin{tabular}{l|cc|cc}
\toprule
     \multirow{2}*{\bf Method} & \multicolumn{4}{c}{\bf ImageNet-50k} \\
     & \bf MACs $\downarrow$ & \bf Latency $\downarrow$ & \bf FID $\downarrow$ & \bf IS $\uparrow$ \\
\midrule
         DiT-XL ($T=20$)~\cite{peebles2023iccv}&  4.579T&  578.201&  15.893
&  \underline{440.797}\\
 DiT-XL ($T=13$)~\cite{peebles2023iccv}& 2.976T& 382.607& 15.982&436.730\\
\midrule 
         
         Faster Diffusion ($I=14$)~\cite{li2023faster}&  3.434T&  458.409&  15.084
&  417.903
\\
         Faster Diffusion ($I=21$)~\cite{li2023faster}&  2.862T&  383.812&  15.145
&  416.609
\\
\midrule 
 $\Delta$-Cache (Front Blocks)& 2.862T& 367.148& 15.112
& 420.198
\\
 $\Delta$-Cache (Middle Blocks)& 2.862T& 368.984& \underline{14.270}& \textbf{442.921}\\
 $\Delta$-Cache (Last Blocks)& 2.862T& 367.042& \underline{13.391}
& 439.700\\
\midrule 
 \rowcolor[HTML]{EFEFEF} 
 Ours ($b=12$)& 2.862T& 370.290& \textbf{13.289}& \underline{442.028}\\
\bottomrule
\end{tabular}}
\label{table:dit-20}
\end{wraptable}
Similar to the results of the PIXART-$\alpha$ architecture, the proposed method outperforms Faster Diffusion and the baseline DiT-XL in FID and IS metrics, with similar or less inference overhead.
In both tables, $\Delta$-Cache shows great results. However, it does not achieve the best in all metrics. For example, on MS-COCO2017, the FID of $\Delta$-Cache (Last Blocks) is the best (34.819), while the CLIP-Score of it (29.898) is lower than most of the other settings. However, $\Delta$-DiT demonstrates outstanding performance across various metrics and datasets.

\subsection{Performance under the Consistent Model}
The consistency model~\cite{song2023consistency, luo2023latent} proposes a new track for generation using few-steps. Essentially, the consistency model can be seen as a refined version, making it highly challenging to accelerate. We evaluated our method in this extreme scenario, with the results shown in Table~\ref{table:pixart-lcm-4}. We found that methods like Faster Diffusion~\cite{li2023faster}, which lack supervision from the previous step images, perform disastrously in small-step scenarios, exhibiting poor image generation metrics (FID=383.812). Existing methods such as TGATE~\cite{zhang2024cross} achieve decent results in this context. However, at an acceleration ratio of approximately 1.12, these methods show significant performance drops (FID: 40.433 $\to$ 44.198). In contrast, our method significantly outperforms this baseline across multiple datasets and metrics in these more extreme conditions. The compression phase of the consistency model's DiT blocks undergoes more pronounced encoding changes, hence a larger $I$ value is chosen, in this case, 8. Furthermore, we can use the FlashEval~\cite{zhao2024flasheval} to quickly evaluate and determine a more appropriate value.

\vspace{-2mm}
\subsection{Ablation Study}
\vspace{-1mm}
\textbf{Compatibility with the advanced solvers.}\ 
The main experiment was conducted using the default solver, DPMSolver++~\cite{lu2022dpm++}. Table~\ref{table:solver} shows the results of our method on several more advanced solvers. It can be seen that the conclusions remain consistent across these solvers, demonstrating that our method is compatible with the current advanced solver. The classic solver DDIM~\cite{song2021ddim} performs poorly in the 20-step PIXART-$\alpha$, so it was not included in the table.

\begin{table}[h]
\centering
\small
\vspace{-4mm}
\caption{Performance under different advanced solvers which are measured on MS-COCO2017.}
\renewcommand{\arraystretch}{1}{
\begin{tabular}{l|ccc|ccc}
\toprule
     \multirow{2}*{\bf Solver}& \multicolumn{3}{c|}{\bf PIXART-$\alpha$}&\multicolumn{3}{c}{\bf +  $\Delta$-DiT}\\
     & \bf FID $\downarrow$& \bf IS $\uparrow$& \bf CLIP $\uparrow$& \bf FID $\downarrow$& \bf IS $\uparrow$&\bf CLIP $\uparrow$\\
\midrule 
 EulerD~\cite{karras2022elucidating}& 39.688& 31.413& 30.359& 35.735& 32.290&30.239
\\
 DEIS~\cite{zhang2023deis} & 37.675& 32.362& 30.420& 35.302& 32.721&30.377
 \\
 DPMSolver++~\cite{lu2022dpm} & 39.002& 31.385& 30.417& 35.882& 32.222&30.404
\\
\bottomrule
\end{tabular}}
\vspace{-2mm}
\label{table:solver}
\end{table}

\begin{figure}[h]
\vspace{-3mm}
    \begin{minipage}[t]{0.33\linewidth}
    \vspace{-4mm}
        \centering
        \small
        \vspace{-75pt}
        \captionof{table}{Results of opposite stage adaptation on PIXART-$\alpha$ and DiT-XL.}
         \resizebox{\linewidth}{!}{ 
        \begin{tabular}{c|ccc}
        \toprule
         \bf Method&  \bf FID $\downarrow$& \bf IS $\uparrow$& \bf CLIP $\uparrow$\\
        \midrule
         \rowcolor[HTML]{EFEFEF} 
         PIXART-$\alpha$-Ours&  35.882&  32.222& 30.404\\
         Opposite&  41.374&  30.980& 30.259\\
         \midrule
          \rowcolor[HTML]{EFEFEF} 
         DiT-XL-Ours&  13.289&  442.028& /\\
        Opposite&  15.255&  426.949& /\\
        \bottomrule
        \end{tabular}}
        \label{tab:ab_oppo}
    \end{minipage}
    \begin{minipage}[t]{0.33\linewidth}
        \centering
        \includegraphics[width=.98\textwidth]{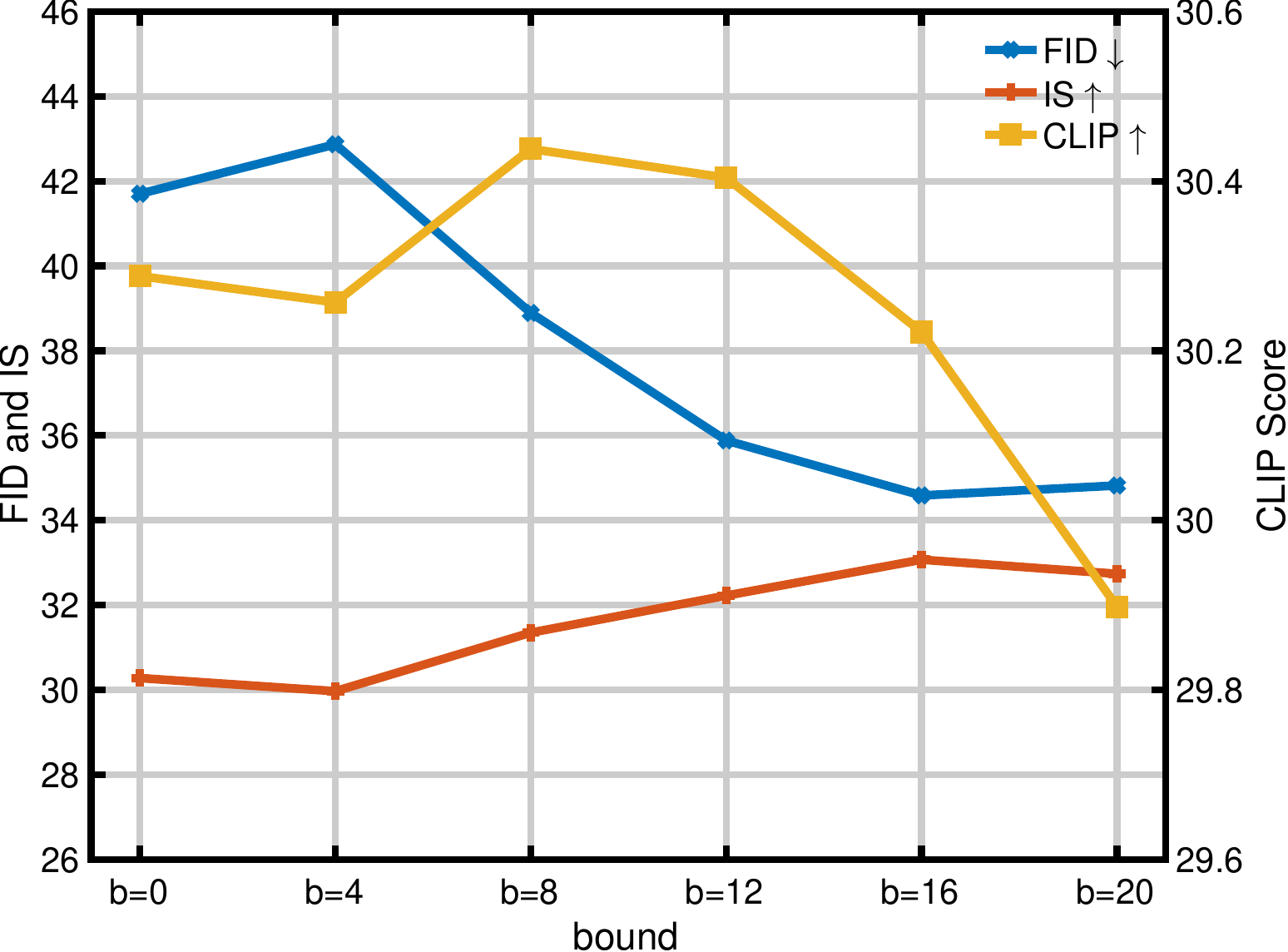}
        \captionsetup{justification=centering} 
        \caption{Ablation study on $b$.}
        \label{fig:ab_bound}
    \end{minipage}%
    \begin{minipage}[t]{0.33\linewidth}
        \centering
        \includegraphics[width=.98\textwidth]{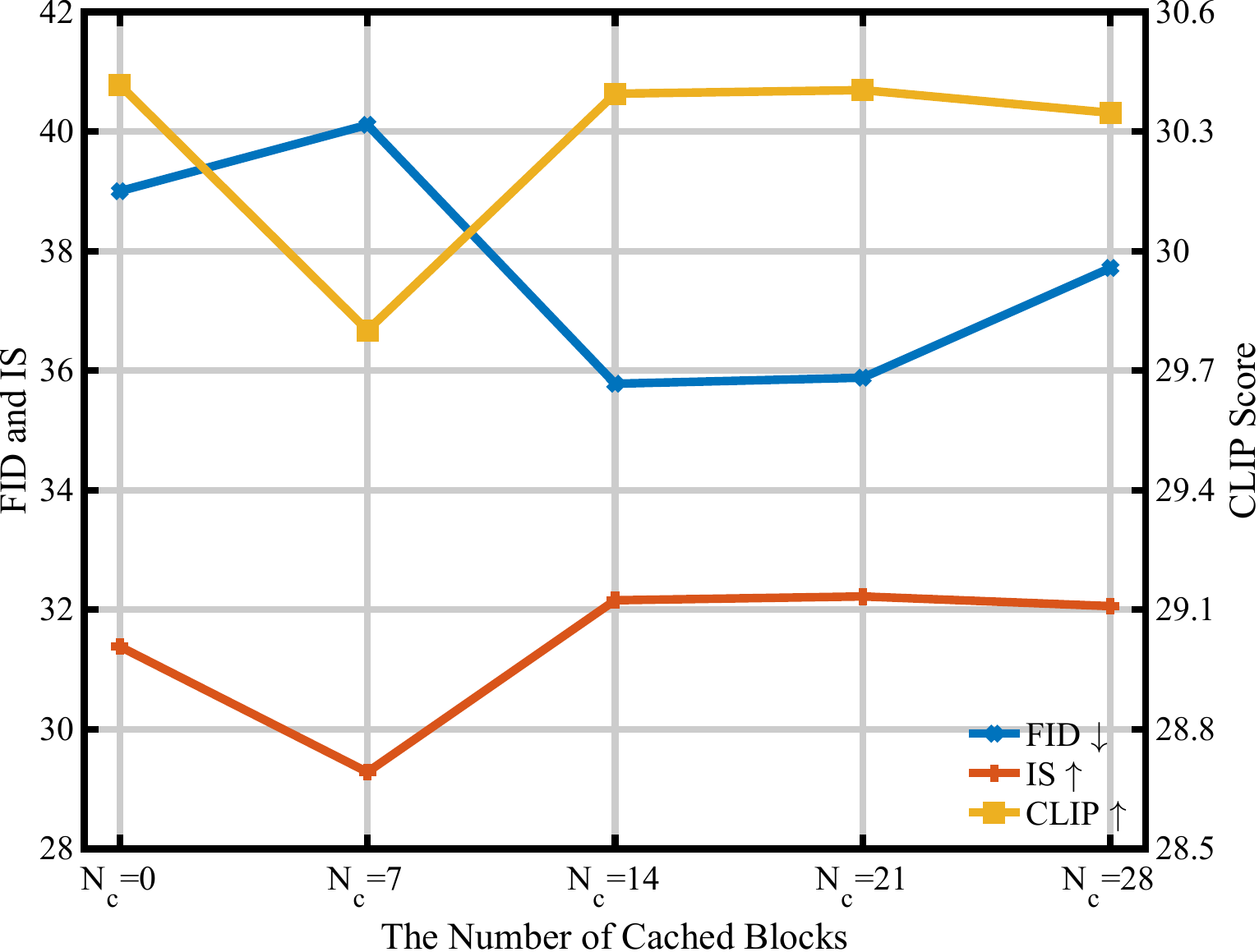}
        \captionsetup{justification=centering} 
        \caption{Ablation study on $N_c$.}
        \label{fig:ab_cache}
    \end{minipage}
    \vspace{-6mm}
\end{figure}

\textbf{Effect of opposite stage adaptation.}\
The $\Delta$-DiT  involves $\Delta$-Cache the back blocks during the early sampling and $\Delta$-Cache the front blocks during the later sampling stages. Here, we reverse the cache order, $\Delta$-Caching the front blocks during the early stages and the back blocks during the later stages. Table~\ref{tab:ab_oppo} compares the results before and after reversing the order. We found that the results significantly deteriorated after reversing the order, which to some extent validates the correctness of our original approach.

\textbf{Illustration of the increasing bound $b$.}\
Figure~\ref{fig:ab_bound} shows the impact of bound on the generation results. As $b$ increases from 0 to 20, FID and IS reach their best values around $b=16$, while the CLIP score peaks around $b=8$. At $b=12$, the three generation metrics are the best overall.

\textbf{Illustration of the increasing number of cached blocks $N_c$.}\ 
Figure~\ref{fig:ab_cache} shows the impact of the number of cached blocks $N_c$ on the generation results. As $N_c$ increases from 0 to 28, FID reaches its optimal value around $N_c=14$, while IS and CLIP Score peak around $N_c=21$. At $N_c=14$ or $21$, the three generation metrics are the best overall.

%% file: latex/conclusion.tex
\section{Conclusion and Limitation}
\vspace{-3mm}
This paper considers the unique structure of DiT and proposes a training-free cache mechanism, $\Delta$-Cache, specifically designed for DiT. Furthermore, we qualitatively and quantitatively explore the relationship between front blocks in DiT and outline generation, as well as rear blocks and detail generation. 
Based on these findings and the sampling properties of diffusion, we propose the stage-adaptive acceleration method, $\Delta$-DiT, which applies $\Delta$-Cache to different part blocks of DiT at different stages of sampling.
Extensive experiments confirm the effectiveness of our approach. However, our exploration of the relationship between DiT blocks and the final generated image is preliminary and coarse-grained, providing a foundation for more fine-grained exploration in future work. We believe that more refined search or learning strategies will yield even greater benefits.